\documentclass{article}
\usepackage{ijcai18}

\usepackage{times}
\usepackage{xcolor}
\usepackage{soul}
\usepackage[utf8]{inputenc}
\usepackage[font=footnotesize]{caption}

\usepackage{savesym}
\usepackage{amsmath,amssymb,amsfonts}
\usepackage{algorithm}
\usepackage[noend]{algpseudocode}
\usepackage{algorithm,algorithmicx}
\usepackage{comment}
\usepackage{xspace}
\usepackage{scalerel}
\usepackage{xcolor}
\usepackage[left]{showlabels}
\usepackage{multicol}
\usepackage{multirow}
\usepackage[bookmarks=true]{hyperref}
\usepackage{graphicx,subfigure}

\title{	Online, Interactive User Guidance for \\
		High-dimensional, Constrained Motion Planning }

\author{
Fahad Islam,
Oren Salzman,
Maxim Likhachev
\\
Robotics Institute, Carnegie Mellon University, USA\\
\{fi,osalzman\}@andrew.cmu.edu,
maxim@cs.cmu.edu
}

\begin{document}

\maketitle
\thispagestyle{empty}
\pagestyle{empty}

\def\frechet{Fr\'echet\xspace}

\newcommand{\cupdot}{\mathbin{\mathaccent\cdot\cup}}

\newcommand{\mtm}{\emph{multi-to-multi}\xspace}
\newcommand{\mts}{\emph{multi-to-single}\xspace}
\newcommand{\sts}{\emph{multi-to-single-restricted}\xspace}
\newcommand{\dtd}{\emph{single-to-single}\xspace}

\newcommand{\cte}{\emph{full-to-edge}\xspace}
\newcommand{\ctc}{\emph{full-to-full}\xspace}
\newcommand{\ete}{\emph{edge-to-edge}\xspace}

\newcommand{\AND}{{\sc and}\xspace}
\newcommand{\OR}{{\sc or}\xspace}

\newcommand{\ignore}[1]{}

\def\vor{\text{Vor}}

\def\P{\mathcal{P}} \def\C{\mathcal{C}} \def\H{\mathcal{H}}
\def\F{\mathcal{F}} \def\U{\mathcal{U}} \def\L{\mathcal{L}}
\def\O{\mathcal{O}} \def\I{\mathcal{I}} \def\E{\mathcal{E}}
\def\S{\mathcal{S}} \def\G{\mathcal{G}} \def\Q{\mathcal{Q}}
\def\I{\mathcal{I}} \def\T{\mathcal{T}} \def\L{\mathcal{L}}
\def\N{\mathcal{N}} \def\V{\mathcal{V}} \def\B{\mathcal{B}}
\def\D{\mathcal{D}} \def\W{\mathcal{W}} \def\R{\mathcal{R}}
\def\M{\mathcal{M}} \def\X{\mathcal{X}} \def\A{\mathcal{A}}
\def\Y{\mathcal{Y}} \def\L{\mathcal{L}}

\def\dS{\mathbb{S}} \def\dT{\mathbb{T}} \def\dC{\mathbb{C}}
\def\dG{\mathbb{G}} \def\dD{\mathbb{D}} \def\dV{\mathbb{V}}
\def\dH{\mathbb{H}} \def\dN{\mathbb{N}} \def\dE{\mathbb{E}}
\def\dR{\mathbb{R}} \def\dM{\mathbb{M}} \def\dm{\mathbb{m}}
\def\dB{\mathbb{B}} \def\dI{\mathbb{I}} \def\dM{\mathbb{M}}

\def\eps{\varepsilon}
\def\obs{\mathrm{obs}}

\newcommand{\sbs}{sampling-based\xspace}
\newcommand{\mr}{multi-robot\xspace}
\newcommand{\mpl}{motion planning\xspace}
\newcommand{\cs}{configuration space\xspace}
\newcommand{\conf}{configuration\xspace}
\newcommand{\confs}{configurations\xspace}
\newcommand{\etal}{et al.\xspace}

\newcommand{\Cpp}{C\raise.08ex\hbox{\tt ++}\xspace}

\newcommand{\ch}{\mathrm{ch}}
\newcommand{\pspace}{{\sc pspace}\xspace}
\newcommand{\np}{{\sc np}\xspace}
\newcommand{\degree}{\ensuremath{^\circ}}
\newcommand{\argmin}{\operatornamewithlimits{argmin}}

\newcommand{\dist}{\textup{dist}}

\newcommand{\Cfree}{\C_{\textup{free}}}
\newcommand{\Cforb}{\C_{\textup{forb}}}

\newtheorem{lemma}{Lemma}
\newtheorem{theorem}{Theorem}
\newtheorem{corollary}{Corollary}
\newtheorem{claim}{Claim}

\newtheorem{definition}{Definition}
\newtheorem{remark}{Remark}
\newtheorem{observation}{Observation}

\def\os#1{\textcolor{blue}{#1}}
\def\ToDo#1{\textcolor{magenta}{\textbf{ToDo:}~#1}}

\makeatletter
\def\thmhead@plain#1#2#3{%
  \thmname{#1}\thmnumber{\@ifnotempty{#1}{ }\@upn{#2}}%
  \thmnote{ {\the\thm@notefont#3}}}
\let\thmhead\thmhead@plain
\makeatother

\def\todo#1{\textcolor{blue}{\textbf{TODO:} #1}}
\def\new#1{\textcolor{magenta}{#1}}
\def\old#1{\textcolor{red}{#1}}

\def\removed#1{\textcolor{green}{#1}}
\algrenewcommand\textproc{}

\newcommand\algname[1]{\textsf{#1}\xspace}
\newcommand\astar{\algname{A*}}
\newcommand\mhastar{\algname{MHA*}}

\newcommand{\arxiv}[2]{#1}
\begin{abstract}
We consider the problem of planning a collision-free path for a high-dimensional robot.
Specifically, we suggest a planning framework where a motion-planning algorithm can obtain guidance from a user.
In contrast to existing approaches that try to speed up planning by incorporating experiences or demonstrations ahead of planning, 
we suggest to seek user guidance only when the planner identifies that it ceases to make significant progress towards the goal.
Guidance is provided in the form of an intermediate configuration~$\hat{q}$, which is used to bias the planner to go through~$\hat{q}$.
We demonstrate our approach for the case where the planning algorithm is Multi-Heuristic \astar (\mhastar) and the robot is a 34-DOF humanoid.
We show that our approach allows to compute highly-constrained paths with little domain knowledge.
Without our approach, solving such problems requires carefully-crafted domain-dependent heuristics. 
\end{abstract}


\section{Introduction}
\label{sec:intro}

Motion-planning is a fundamental problem in robotics that has been studied for over four \arxiv{decades~\cite{CBHKKLT05,L06,HSS17}}{decades~\cite{L06,HSS17}}.
However, efficiently planning paths in high-dimensional, constrained spaces remains an ongoing challenge.
One approach to address this challenge is to incorporate user input to guide the motion-planning algorithm.
While there has been much work on planning using human demonstration 
\arxiv{(see, e.g.,~\cite{ACVB09,PHCL16,SHLA16,YA17}),}{(see, e.g.,~\cite{ACVB09,YA17}),}  
there has been far less research on incorporating guidance as an interactive part of the planning~loop.

Broadly speaking, interactive planning has been typically used in the context of sampling-based motion-planning algorithms~\cite{L06}.
User guidance was employed for biasing the sampling scheme of the planner by having the user mark regions in the \emph{workspace} that should be avoided or 
\arxiv{explored~\cite{DSJA14,MTMKDC15,YPB15,RSL18}.}{explored~\cite{DSJA14,YPB15,RSL18}.}
Alternatively, interactive devices such as~a 3D mouse or a haptic arm have been used to generate paths in~a (low-dimensional) configuration space. Such paths were then used by a planner to bias its 
\arxiv{sampling domain~\cite{BTFF16,FTF09,TFF12}.}{sampling domain~\cite{TFF12}.}

\begin{figure}[tb]
  \centering
  	\includegraphics[width=0.4\textwidth]{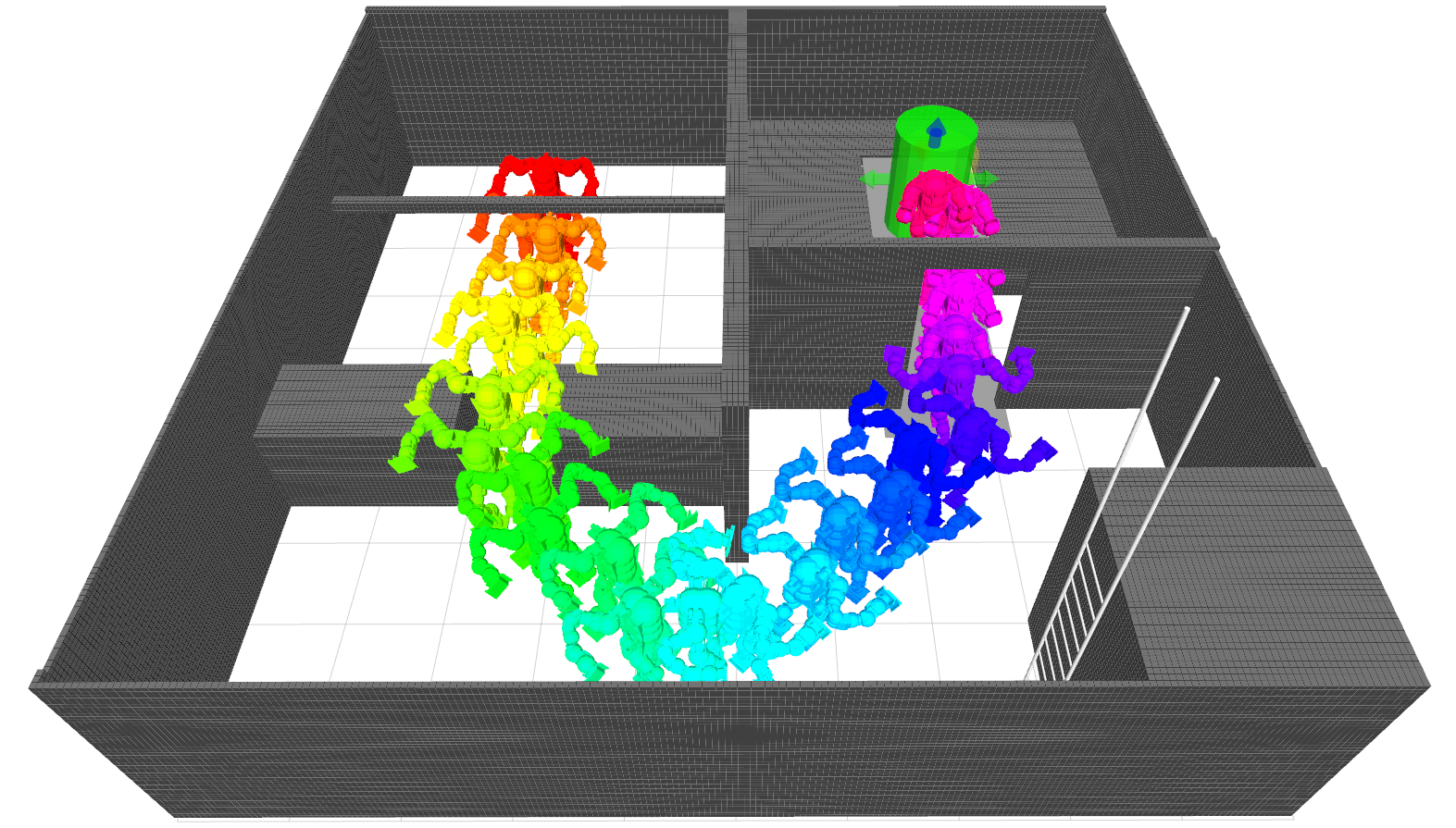}
  \caption{
  Planning domain---Humanoid robot needs to locomote in a challenging environment that involves circumventing obstacles and passing through narrow passages. Shown in multi-color are sparsed states from the path generated by our proposed approach.
}
   	\label{fig:robot}
\end{figure}

We are interested in planning in high-dimensional, constrained spaces such as those encountered by a humanoid robot (see Fig.~\ref{fig:robot} and Sec.~\ref{sec:rel}).
In such settings, workspace regions often give little guidance to the planner due to the high dimension of the configuration space as well as the physical constraints of the robot.
Additionally, providing the user guidance in the configuration space is extremely time consuming, even for expert users.
Thus,  though beneficial, user guidance should be employed scarcely.

Our key insight is that carefully chosen individual configurations suggested by a user can be used to effectively guide the planner when in need.
Transforming this insight into a planning framework requires addressing three fundamental questions:

\begin{itemize}
	\item[\textbf{Q1.}] When should the planner ask the user for guidance?
	\item[\textbf{Q2.}] What form should the user's guidance take?
	\item[\textbf{Q3.}] How should the guidance be used by the planner?
\end{itemize}

Identifying \emph{when} to obtain guidance 
(Q1, Sec~\ref{sec:q1}) 
comes in stark contrast to existing approaches---we suggest to only employ user guidance when the planner \emph{identifies} that it ceases to make significant progress towards the goal.
Once obtained (Q2, Sec~\ref{sec:q2}), guidance is used to \emph{bias} the search algorithm towards regions that are likely to be beneficial (Q3, Sec~\ref{sec:q2}). 
The search algorithm now has the additional task of deducing how much to rely on the user-provided guidance.
It is worth emphasizing that obtaining this guidance does not require the user to understand the underlying search algorithm. 

%

%
We demonstrate our approach using search-based planning algorithms (see, e.g.,~\cite{CCL14}) that perform a systematic search guided by heuristic functions.
Specifically, we use multi-heuristic \astar (\mhastar)~\cite{ASNHL16,NAL15} which we detail in Sec.~\ref{sec:mha}.
Our conjecture is that the formulation can, in general, be incorporated with any motion-planning algorithm if it can address the aforementioned questions, however showing its effectiveness on other algorithms is out of the scope of the paper.

%

After describing our approach at high-level (Sec.~\ref{sec:high}) we demonstrate how it can be applied to the case of \mhastar (Sec.~\ref{sec:planning}).
We continue by showing the effectiveness of our planner (Sec.~\ref{sec:eval}).
Specifically, we show that this general approach allows to compute highly-constrained motion plans for a spectrum of tasks including climbing stairs, walking under a bar, squeezing through a door, and others, all with the same planner and heuristics.  
Without our approach, solving such problems requires the design of task-specific planners~\cite{KKKHKHAI04} and carefully hand-designed domain-dependent heuristics.

\section{Related Work and Algorithmic Background}
\subsection{Motion Planning for Humanoid Robots}
\label{sec:rel}
Humanoid robots, for which we demonstrate our approach, often have several dozens of degrees of freedom making planning a challenging task, especially when taking additional constraints into account such as stability and contact constraints.
One approach to plan the motion for such systems, is to use predefined, carefully chosen fixed gaits~\cite{KKKHKHAI04}. 
However, when the terrain is uneven, such planners are inadequate at computing stable motions~\cite{HBLHW08}.
Another approach is to reduce the size of the search space by decomposing the degrees of freedom into functional groups such as locomotion and manipulation.
Then functional-specific algorithms such as footstep planning are applied to the low-dimensional space 
\arxiv{(see, e.g.,~\cite{CLCKHK05,PSBLY12,XCXZC09%
} for a partial list).}{(see, e.g.,~\cite{PSBLY12,XCXZC09} for a partial list).}
A high-dimensional planner is then used to ``track'' the plan generated in the low-dimensional space.

User guidance has been intensively incorporated in controlling the motion of humanoid robots, especially in complex tasks as those presented in the Darpa Robotics Challenge (DRC)~\cite{drc18}.
In such settings, guidance ranged from teleoperating the robot 
to high-level 
\arxiv{task guidance~\cite{mcgill2017team,%
gray2017architecture,dedonato2017team,marion2017director}.}{{task guidance~(see e.g.,~\cite{mcgill2017team,gray2017architecture}).}}
However, as far as we are aware of, in all the aforementioned cases, \emph{identification} of when to use guidance was done by a human operator and not by the system (except for relatively simple metrics where the system asks for help when it fails to complete some given task).
Furthermore, the human guidance was used as a ``hard'' constraint forcing the system to make use of the guidance.
In contrast, in our work the system automatically and independently identifies when it is in need of guidance.
This guidance is then seamlessly incorporated as a soft constraint--- the system biases the search towards the guidance while continuing to explore the search space as if the guidance was not given.

\subsection{Multi Heuristic \astar (\mhastar)}
\label{sec:mha}
Multi Heuristic \astar (\mhastar)~\cite{NAL15,ASNHL16} is a search-based planning algorithm that takes in multiple, possibly inadmissible heuristic functions in addition to a single consistent heuristic termed the \emph{anchor} heuristic.
It then uses the heuristic functions to simultaneously perform a set of weighted-\astar~\cite{pohl1970first}-like searches.
Using multiple searches allows the algorithm to efficiently combine the guiding powers of the different heuristic functions. 

Specifically, for each search, \mhastar uses a separate priority queue associated with each heuristic. 
The algorithm iterates between the searches in a structured manner that ensures bounds on sub-optimality. 
This can be done in a round-robin fashion, or using more 
sophisticated approaches that allow to automatically calibrate the weight given to each heuristic~\cite{PNAL15}.

Part of the efficiency of \mhastar is due to the fact that the value of the cost-to-come (the $g$-value) computed for each state is shared between all the different searches\footnote{To be precise, there are two variants of \mhastar described in~\cite{ASNHL16}: Independent and Shared \mhastar where the queues do not share and do share the $g$-values of states, respectively. In this paper when we use the term \mhastar, it refers to the latter (shared) variant.}.
Sharing cost-to-come values between searches implies that if a better path to a state is discovered by any of the searches, the information is updated in all the
priority queues. 
Moreover, if one search ceases to make progress towards the goal, it can use ``promising'' states visited by other searches to escape the local minimum.

\subsection{Identifying Progress in Search-based Planners}

A key component in our work is automatically detecting when our planner requires user guidance.
This requires \emph{characterizing} regions where the planner ceases to make progress and algorithmic tools to \emph{detect} such regions. 
These two requirements are closely related to the notion of~\emph{heuristic depressions}~\cite{I92}
and~\emph{search vacillation}~\cite{DTR11,BRD13}.

A heuristic depression region is a  region in the search space where the correlation between the heuristic values and the actual cost-to-go is weak.
It is defined as a maximal connected component of states~$\mathcal{D}$ such that all states in the boundary of~$\mathcal{D}$ have a heuristic value that is greater than or equal to the heuristic value of any state in~$\mathcal{D}$.

Such regions often occur in real-time search algorithms such as \algname{LRTA*}~\cite{K90} and \algname{LSS-LRTA*}~\cite{KS09} where the heuristic function is updated as the search progresses.
Subsequently, current state-of-the-art algorithms  guide the search to avoid states that have been marked as part of a heuristic depression~\cite{HB12}.
As we will see, we will use a slightly different characterization of when the planner ceases to make progress which we call \emph{stagnation regions}. For details, see Sec.~\ref{sec:q1}.

Search vacillation is a phenonemon when the search explores multiple small paths due to error in the heuristic. 
We will use the notion of search vacillation to identify when the planner is in a stagnation region (i.e., when it ceases to make progress towards the goal). For details, see Sec.~\ref{sec:q1}.

\section{Algorithmic Approach---User-guided Planning}
\label{sec:high}

\algrenewcommand\algorithmicindent{.8em}
\begin{algorithm}[tb]
\caption{User-guided Planning ($\A$)}
\label{alg:main}	
\begin{algorithmic}[1]
\small
\While{$\neg\A.$\texttt{solution\_found()} } 
	\While{$\neg\A.$\texttt{in\_stagnation\_region()}} 
		\State $\A.$\texttt{run()}
		\Comment{no user guidance}
	\EndWhile
	\State {$g \leftarrow$ \texttt{get\_user\_guidance()}}
	\Comment{$\A$ in a stagnation region}
	\State $\A.$\texttt{update\_user\_guidance($g$)}
	\Comment{account for guidance}
	\While{$\A.$\texttt{in\_stagnation\_region()} \textbf{and} $\neg\A.$\texttt{in\_stagnation\_region($g$)}}
		\State $\A.$\texttt{run()}
		\Comment{$\A$ uses guidance to escape stagnation region}
	\EndWhile

	\State $\A.$\texttt{update\_user\_guidance($\neg g$)}
	\Comment {remove  guidance}
\EndWhile

\end{algorithmic}
\end{algorithm}

To employ our planning framework, we assume that we are given a motion-planning algorithm $\A$ that is endowed with two non-standard procedures which are planner dependent.
The first, \texttt{in\_stagnation\_region()}, 
identifies when it is in a \emph{stagnation region} (with or without the guidance), namely when $\A$'s search does not progress towards the goal. 
The second, \texttt{update\_user\_guidance()}, 
incorporates (or removes) the user guidance provided to $\A$. 

Equipped with these functions, we can describe our planning framework, detailed in Alg.~\ref{alg:main}.
The framework runs as long as no solution is found (line~1).
It runs the planner~$\A$ (lines~2-3) as long as it continuously makes progress towards the goal (namely, it is not in a stagnation region).
Once a stagnation region is identified, user guidance is invoked (line~4) and $\A$  is updated to make use of this guidance (line~5).
The planner continues to run while using the guidance as long as the planner without the guidance is still in the stagnation region, and the guidance is useful (lines~6-7).
Once it escapes the stagnation region (or if the guidance appears to be unhelpful), $\A$ is updated to remove the guidance that was provided by the user (line~8).

\section{User-guided Planning via \mhastar}
\label{sec:planning}

We demonstrate our general planning framework described in Sec.~\ref{sec:high} for the case where the motion-planning algorithm~$\A$ is  \mhastar.
We assume that \mhastar has a set of possibly inadmissible heuristics in addition to the anchor heuristic. We will refer to these heuristics as \emph{baseline heuristics}.
The user guidance will be used to generate \emph{dynamic heuristics}.
In Sec.~\ref{sec:q1}-\ref{sec:q3} we describe how we answer the three questions that were posed in Sec.~\ref{sec:intro}.

\subsection{Invoking User Guidance (Q1)}
\label{sec:q1}

\begin{figure}
  \centering
  \includegraphics[width=0.35\textwidth]{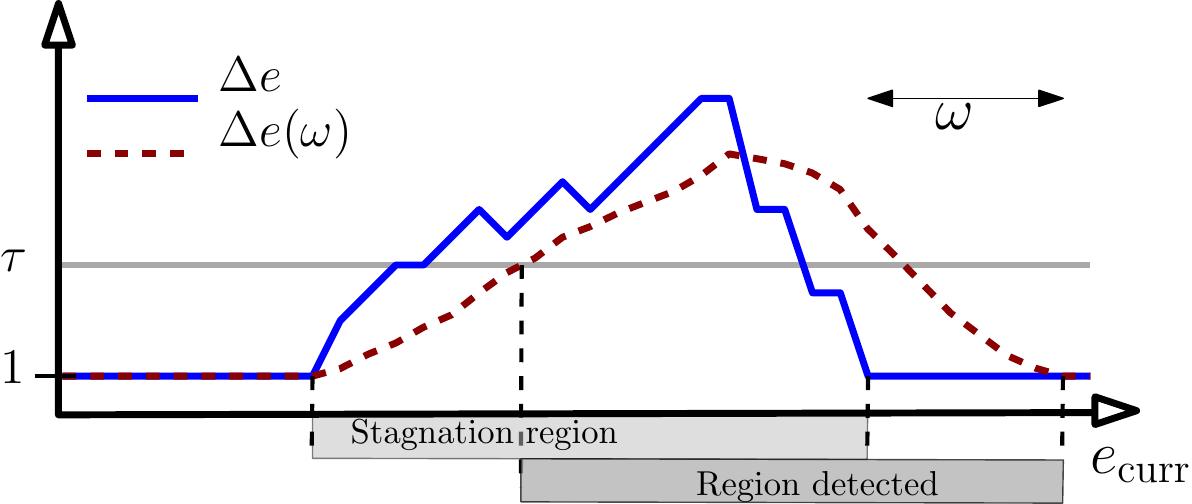}
	\vspace{-2mm}

  \caption{%
    Visualization of the way vacillation-based stagnation regions are detected---expansion delay as a function of the number of expansions.  
    Notice that the stagnation regions ends when $\Delta e = 1$.
    However, this is detected after $\omega$ additional steps.}

	\vspace{-1.5mm}
  \label{fig:filmstrip-local-min}%
\end{figure}

\begin{figure*}[t]%
  \centering%
  \subfigure[]
  {
  \includegraphics[width=0.324\textwidth]{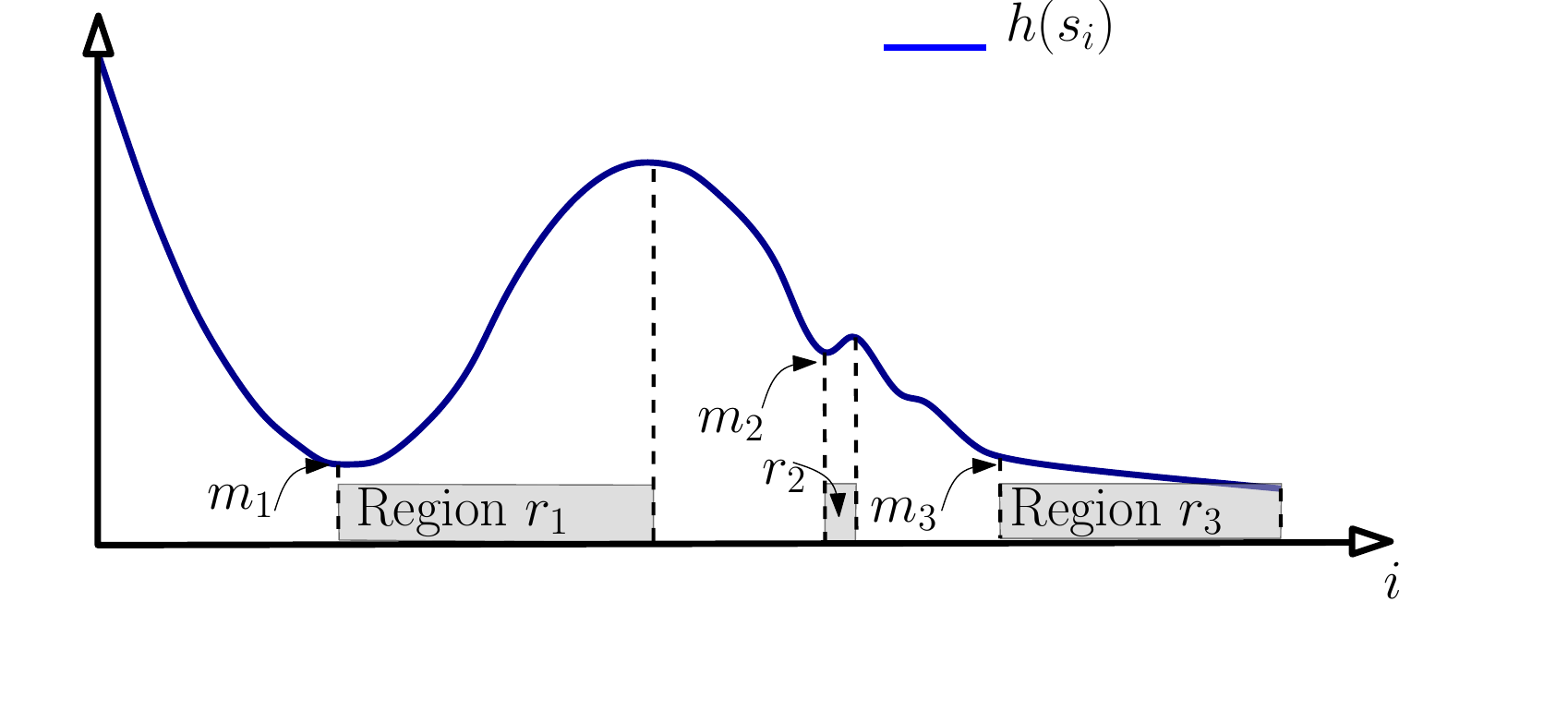}
  \label{fig:local_min1}
  }
  \hspace{-5mm}
  \subfigure[]
  {
  \label{fig:local_min3}
  \includegraphics[width=0.324\textwidth]{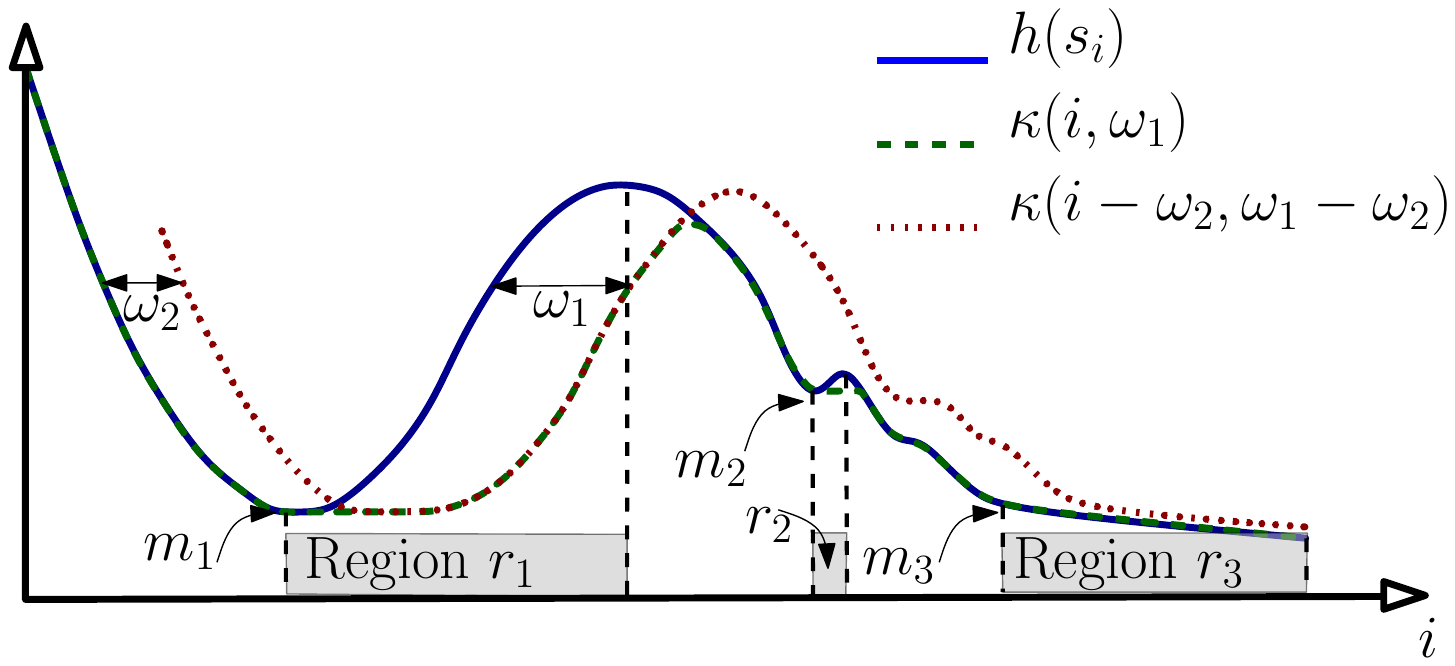}
  }  
  \hspace{-5mm}
  \subfigure[]
  {
  \label{fig:local_min4}
  \includegraphics[width=0.324\textwidth]{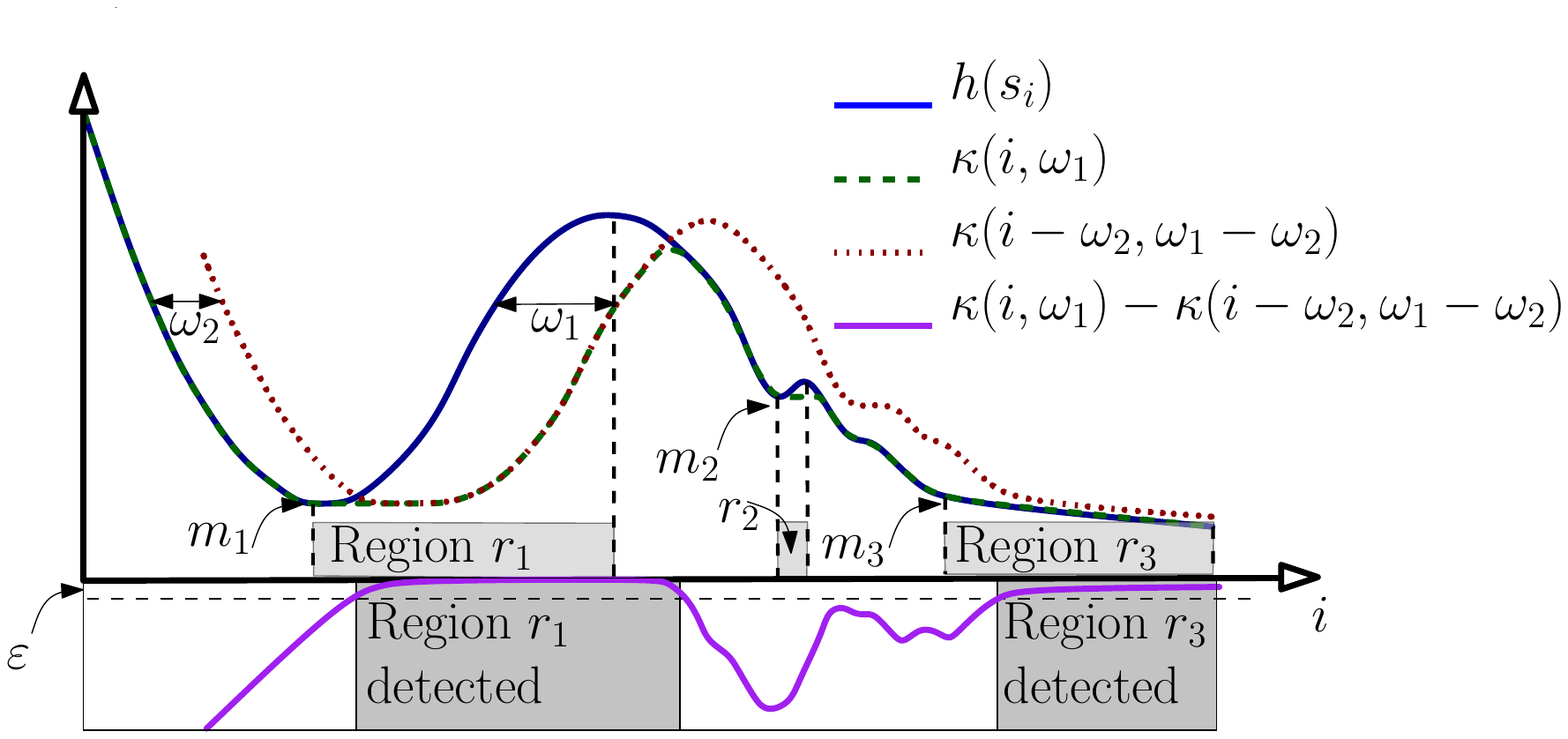}
  }
  \vspace{-2mm}
  \caption{%
    Visualization of heuristic-based stagnation-region detected.   
	\subref{fig:local_min1}~Heuristic $h(s_i)$ (solid blue) has three local minima, $m_1, m_2$ and $m_3$ followed by three stagnation regions (light grey). Local minimum $m_2$ is very small while~$m_3$ is not a local minimum per se, yet the progress made between consecutive steps is smaller than the predefined threshold~$\varepsilon$.
    \subref{fig:local_min3}~The function $\kappa(i,\omega_1)$ (dashed green) returns the minimal value $h(s_i)$ attained over the past~$\omega_1$ iterations.
    %
    The function $\kappa(i-\omega_2,\omega_1-\omega_2)$ (dotted red) returns the minimal value $h(s_i)$ attained over the past ~$\omega_1$ iterations excluding the last~$\omega_2$ iterations.
    \subref{fig:local_min4}~The difference between the two functions (solid purple) indicates if there was significant progress (i.e., more than $\varepsilon$) made over the last~$\omega_2$ iterations. 
    If not, then the planner detects a stagnation region (dark grey).
		Notice that the hysteresis parameters~$\omega_1$ and~$\omega_2$ 
		(i)~induce a lag from the time that the stagnation region starts until it is detected,
		(ii)~allow to avoid detecting stagnation region~$r_2$.  
		}%

  \label{fig:filmstrip-local-min2}%


\end{figure*}

The heuristic functions of search-based planning algorithms, such as \mhastar, can be used to estimate in a principled manner when the planner is in a stagnation region (Alg~\ref{alg:main}, lines 2 and~6). 
We suggest two alternative methods to identify when the planner ceases to make progress towards the goal, each resulting in a slightly different definition of a  stagnation region.
The first, which we call \emph{vacillation-based} detection uses the notion of expansion delays~\cite{DTR11}:
Let~$\Q$ be a priority queue and let $e_\text{curr}$ be a counter tracking the total number of node expansions performed when using~$\Q$.
When a node is expanded, for each child node $s$ we store the current expansion number using a counter $e(s)$.
Now, the expansion delay for a state $s$ is defined as
$$
\Delta e = e_\text{curr} - e(s).
$$ 

A moving average of the expansion delay of the nodes being expanded can be used as a proxy to measure the average progress that a planner makes along any single path.
When the heuristic used by a planner is perfect, the expansion delay equals one, while on the other extreme, when performing uniform-cost search, the expansion delay can grow exponentially.

Given some parameter $\omega >0$ we compute $\Delta e(\omega)$, the average expansion delay over a moving window of size $\omega$. 
\begin{definition}
\label{def:expansion_delay}
Let 
$\omega>0$ be some window size
and
$\tau$ be some threshold value.
A heuristic~$h$ associated with a queue~$\Q$ is defined to
\begin{enumerate}
	\item enter a stagnation region if $\Delta e(\omega) \geq \tau$,
	\item exit a stagnation region if $\Delta e(\omega) = 1$
\end{enumerate}
\end{definition}
For a visualization of vacillation-based stagnation regions and how they are detected (Def.~\ref{def:expansion_delay}), see Fig.~\ref{fig:filmstrip-local-min}.

As we will see in our experimental section, due to the nature of the domain or the heuristic function, vacillation-based detection could be deceptive and not detect regions where the planner is not making significant progress. 
Thus, we suggest an alternative method, which we call \emph{heuristic-based} detection.
Let~$\Q$ be a priority queue 
ordered according to some heuristic function~$h(\cdot)$,
$s_i$ be the node expanded from~$\Q$ at the $i'$th iteration and $\omega_1, \omega_2$ and $\varepsilon$ be parameters such that $\omega_1 > \omega_2$ and $\varepsilon$ is some threshold.
We define 
$\kappa_\Q(i, \omega) = \min_{i-\omega \leq j \leq i} \{ h(s_j)\}$.
Namely, $\kappa_\Q(i, \omega)$ denotes the minimal value attained by $h$ over the past ~$\omega$ expansions. 
\begin{definition}
\label{def:heur}
A heuristic~$h$ associated with a queue~$\Q$ is defined to be in a (heuristic-based) stagnation region if 
$\kappa_\Q(i, \omega_1) \geq \kappa_\Q(i - \omega_2, \omega_1 - \omega_2) - \varepsilon$.
\end{definition}
\noindent Namely,~$\Q$ is in a heuristic-based stagnation region if looking at the previous~$\omega_1$ iterations, 
there was no reduction 
(by more than $\varepsilon$) 
in the minimum value of~$h$ 
in the last~$\omega_2$ states expanded from $\Q$.
%
%
For a visualization of heuristic-based stagnation regions and how they are detected (Def.~\ref{def:heur}), see Fig.~\ref{fig:filmstrip-local-min2}.

\subsection{Form of User Guidance (Q2)}
\label{sec:q2}
\begin{figure}[tb]
  \centering
  	\includegraphics[width=0.46\textwidth]{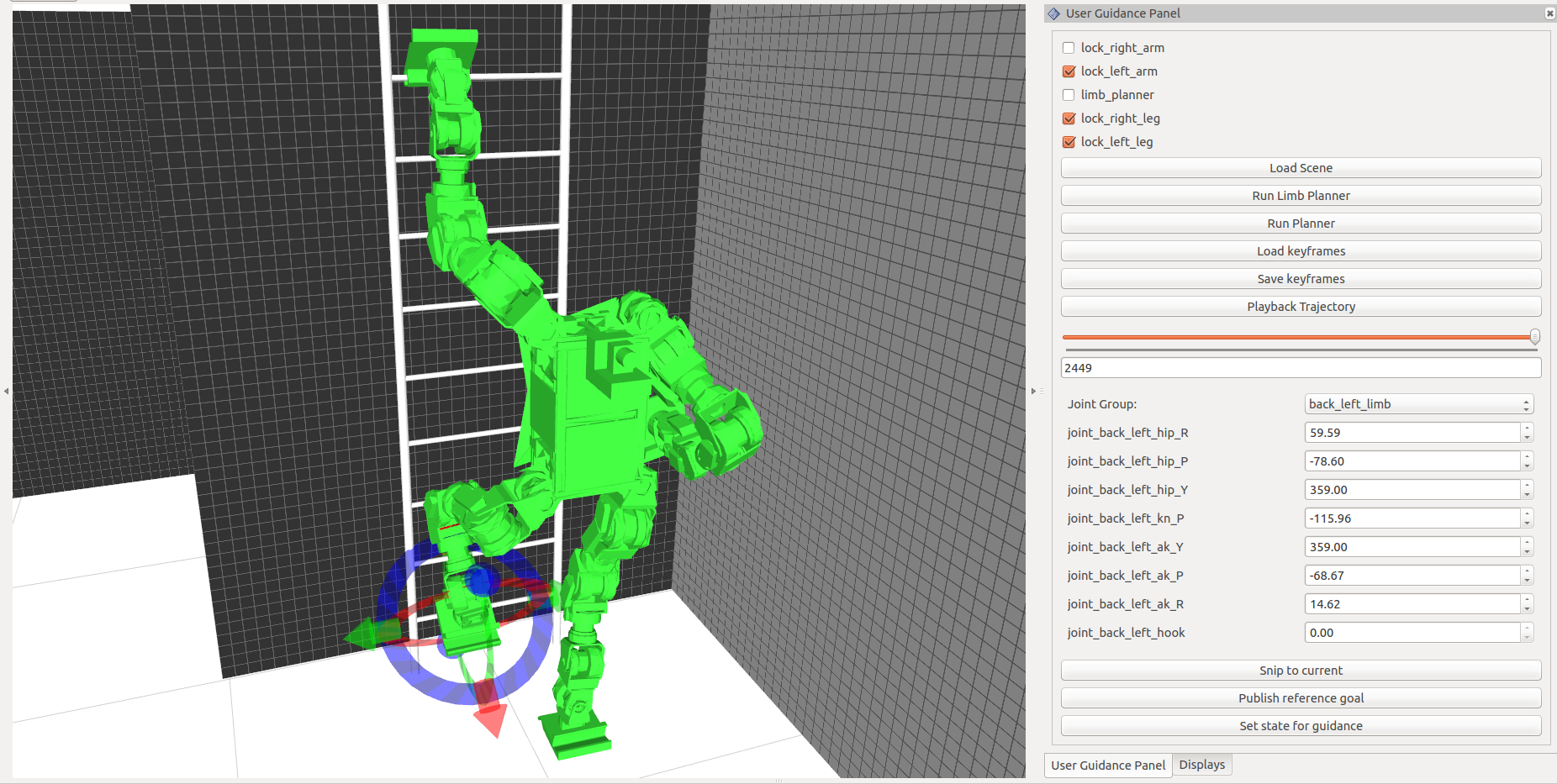}
  \caption{
		GUI used to provide guidance to the planner. The panel on the right hand side allows to select different joint groups, move the joints and pass the guidance to the planner. The 6 DOF interactive marker shown on the left hand side allows to move the robot's end effectors and its torso in the task space. The user can move the torso while having specified end-effectors locked in place. The tool runs an inverse-kinematics solver to enable such movements.
}
   	\label{fig:gui}
  	\vspace{-2mm}
\end{figure}

We chose to obtain user guidance 
(Alg~\ref{alg:main}, line~4)
in the form of an intermediate configuration $\hat{q}$ that is used to guide the planner. 
The framework includes a graphical user interface (GUI) (Fig.~\ref{fig:gui}) capable of  depicting the robot and the workspace.
Once user guidance is invoked, 
a configuration in the stagnation region is obtained and the robot is placed in that configuration in the GUI.
This allows the user to intuitively try and understand where the planner faces difficulty and how to guide it out of the stagnation region.
The user then suggests the guidance $\hat{q}$ by moving the robot's joints, end effectors or torso.
The tool runs a state validity checker (the same one as used by the planner during the search) in the background which restricts the user from providing invalid configurations, e.g. with respect to collision, joint limits and stability.

We informally experimented with non-expert users, and after some practice with the GUI received positive feedback. Our focus in this work was on the algorithmic aspects of our approach and we leave UI challenges to future work.

\subsection{Using User Guidance (Q3)}
\label{sec:q3}

\begin{figure*}[t]%
  \centering%
  \subfigure[]
  {
  \includegraphics[width=0.182\textwidth]{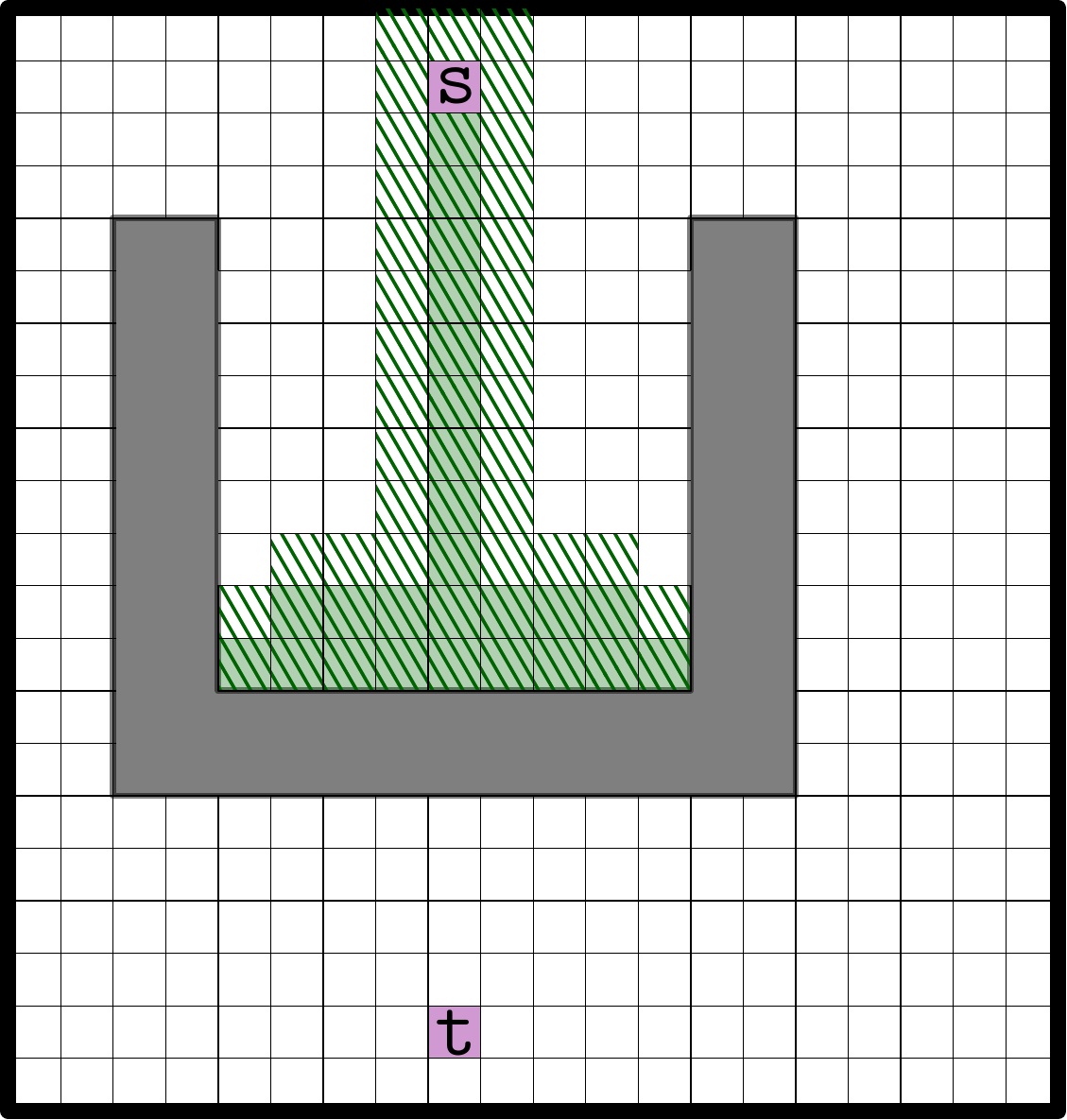}
  \label{fig:dynamic_heuristic1}
  }
  \subfigure[]
  {
  \label{fig:dynamic_heuristic2}
  \includegraphics[width=0.182\textwidth]{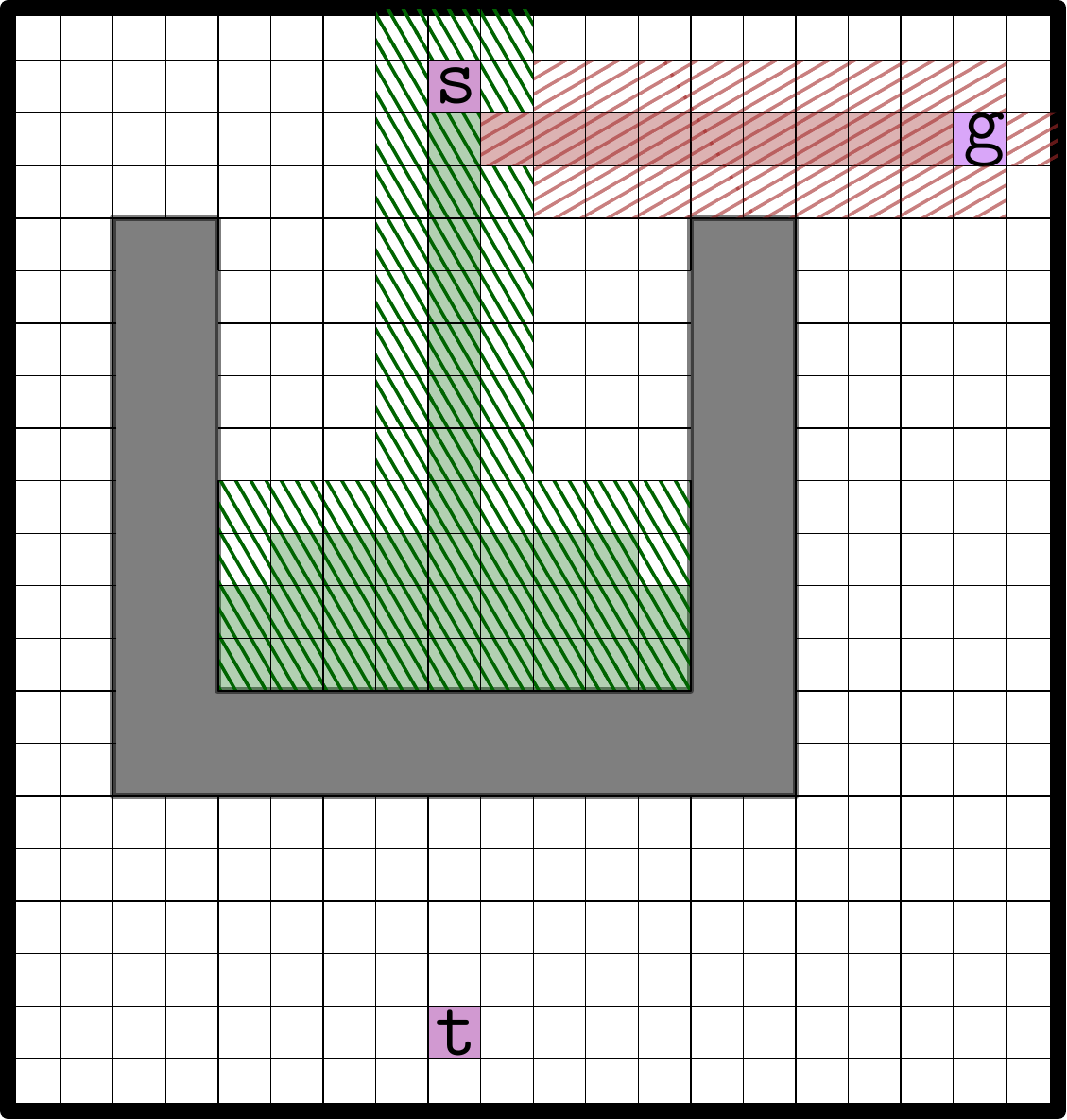}
  }
  \subfigure[]
  {
  \label{fig:dynamic_heuristic3}
  \includegraphics[width=0.182\textwidth]{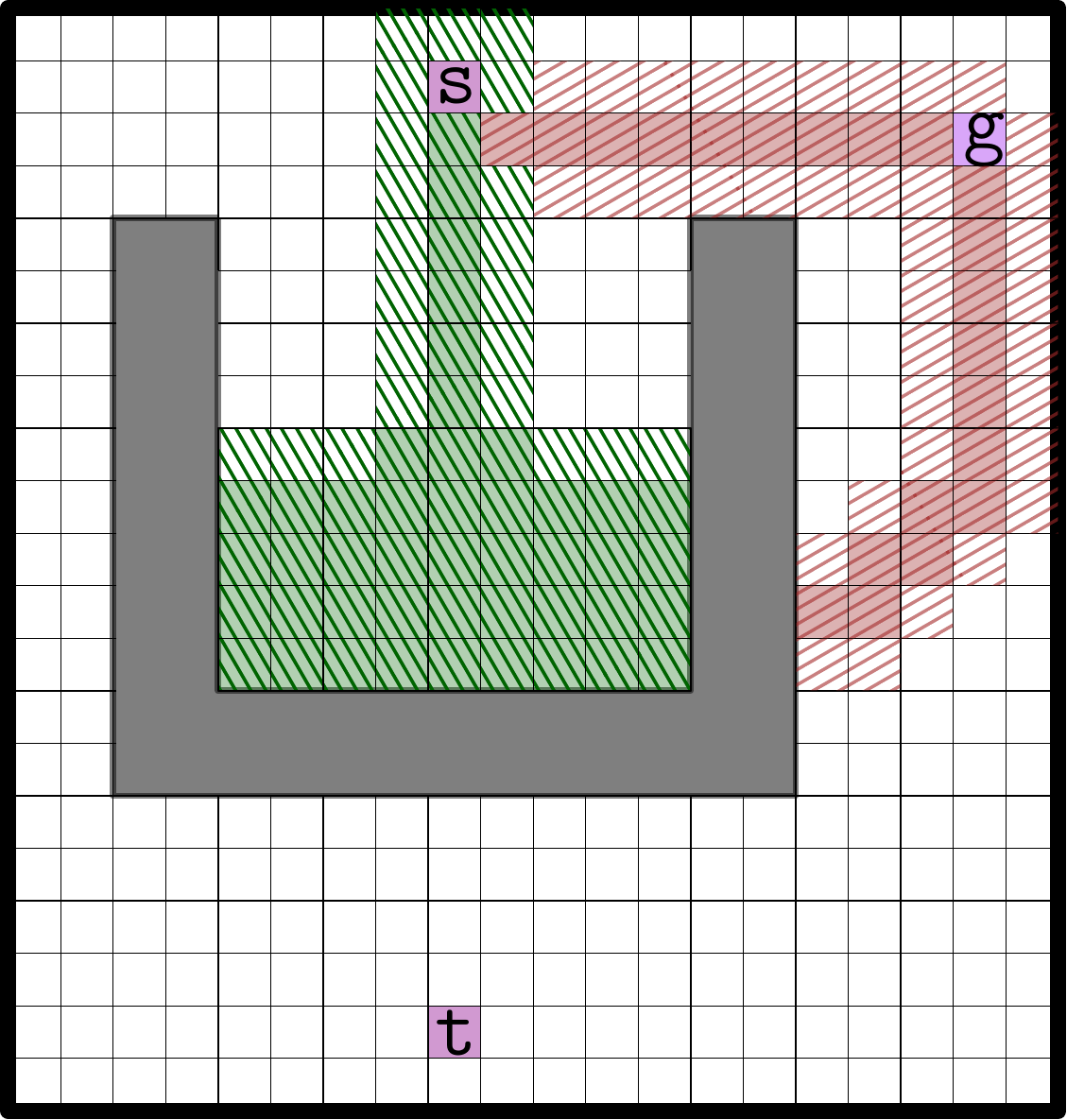}
  }  
  \subfigure[]
  {
  \label{fig:dynamic_heuristic4}
  \includegraphics[width=0.182\textwidth]{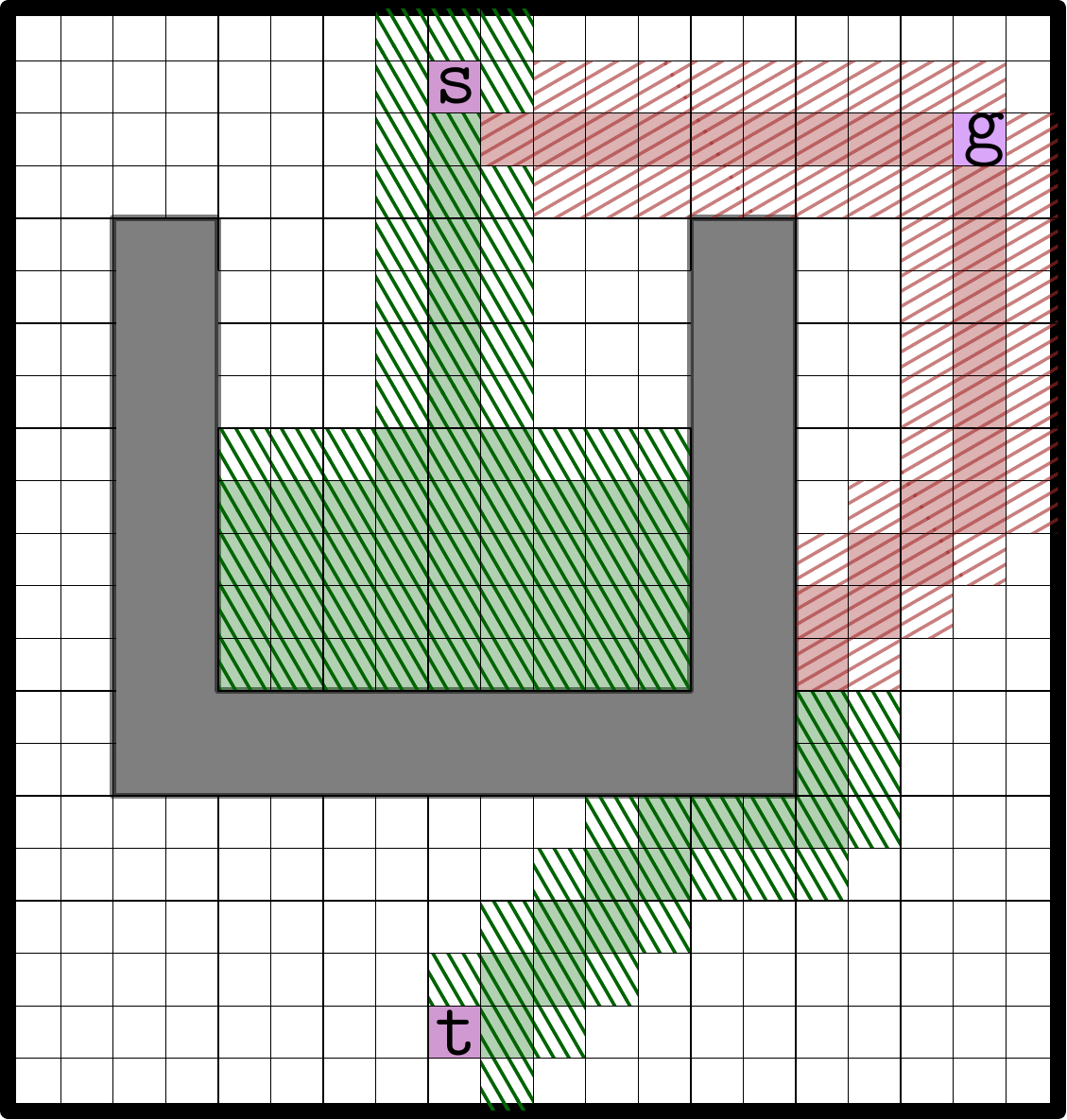}
  }
  \caption{%
    Algorithm progression.
    States popped from a priority queue~$Q$ and those that are still in~$Q$ are depicted using dark and light colors respectively.
    Start, target and user-guided states are depicted in purple with the letters \texttt{s}, \texttt{t}, \texttt{g}, respectively. 
    In this example \mhastar alternates between queues in a round-robin fashion and heuristic values are inflated by a weight of $w=\infty$.
    Namely, each search greedily follows its heuristic and pops the state with the minimal heuristic value.
    %
	\subref{fig:dynamic_heuristic1}~\mhastar starts with a single baseline heuristic (green, falling pattern) which is the Euclidean distance to goal and a stagnation region is identified.
	\subref{fig:dynamic_heuristic2}~User provides guidance~$\texttt{g}$ and an additional heuristic (red, rising pattern) is automatically generated and drives the search towards the guidance (notice that the baseline heuristic continues to search and that the second, red search, starts from the green state that is closest to the guidance).
	\subref{fig:dynamic_heuristic3}~After passing through~$\texttt{g}$, the additional heuristic (red) drives the search towards the goal.
	Notice that all the green states have lower Euclidian distance to  the target, meaning that the green search continues to expand states in the U-shaped obstacle.
	\subref{fig:dynamic_heuristic4}~After the additional heuristic found states that are placed at the top of the priority queue of the baseline (green) search, the additional heuristic is deleted and the baseline heuristic continues to drive the search towards the goal.
  }%
  \label{fig:filmstrip-dynamic_heuristic}%
\end{figure*}

We assume that \mhastar has at least one baseline (possibly inadmissible) heuristic~$h_{\text{goal}}$ (in addition to the anchor heuristic) which approximates the cost to reach the goal from every state.
Furthermore, we assume that for every configuration~$q$, there exists a heuristic $h_q$ where~$h_q(s)$ estimates the cost to reach $q$ from state $s$.
%

Given user guidance in the form of a configuration $\hat{q}$, we dynamically generate a new heuristic $$
    \hat{h}(s)= 
\begin{cases}
    h_{\hat{q}}(s) + h_{\text{goal}}(\hat{q}),	& 
    		\text{if } \hat{q} \text{ is not an ancestor of } s,\\
    h_{{goal}}(s),            		& 
    		\text{if } \hat{q} \text{ is an ancestor of } s.
\end{cases}
$$
Namely,~$\hat{h}$ estimates the 
cost to reach the goal via $\hat{q}$ (see also~\cite{CGD86} for intuition regarding such a heuristic). 
If the state was reached by passing through~$\hat{q}$, then the value of $\hat{h}$ is simply the estimation of the cost to reach the goal.

Equipped with the heuristic $\hat{h}$, we add a new queue to \mhastar prioritized using the~$\hat{h}$. 
States expanded using this queue will be biased towards $\hat{q}$ (see also~\cite{INL15} for more details on adding heuristics dynamically to \mhastar
).
Recall that in \mhastar, nodes are shared between the different queues.
Thus, once a state has been found that can be used to get the planner out of the stagnation region, it will be expanded by the other queues using their heuristics.
As the planner escapes the stagnation region,
the newly-added queue is removed.
In general,  we can add a dynamic queue for every baseline heuristic if they are more than one. However, for simplicity, in this work we use a single baseline heuristic and add one dynamically generated queue when the user provides guidance. Since the search runs under the \mhastar framework, although the dynamic heuristic $\hat{q}$ is inadmissible, we still maintain completeness and bounded sub-optimality guarantees.

For pseudo-code describing each specific function used  in Alg.~\ref{alg:main} under the framework of \mhastar, see Alg.~\ref{alg:instantiation}.
When adding the dynamic heuristic, if the baseline heuristic escaped a stagnation region but the configuration $\hat{q}$ was \emph{not} reached, we suspend the dynamic queue but do not discard it. 
This is done to first try reusing the last guidance before obtaining a new one. 
Thus, when the planner will detect that it is in a stagnation region, it will first resume the suspended dynamic heuristic (if one exists).
If the baseline heuristic escaped a stagnation region and the configuration~$\hat{q}$ was  reached it will no longer be useful again and hence will be discarded.
Finally, if the dynamic heuristic is in a stagnation region then it is discarded and the user will be queried for a new guidance. 
We emphasize that detecting stagnation is done for \emph{each} heuristic independently.

For a visualization of the way the algorithm progresses, see Fig.~\ref{fig:filmstrip-dynamic_heuristic}. Note that although for the illustrated example, the search happens to pass through the guidance, in general it is not a constraint in the framework. 
%

\begin{algorithm}[tb]
\caption{User-guided \mhastar}
\label{alg:instantiation}	
\begin{algorithmic}[1]
\small
\Function{\texttt{in\_stagnation\_region()}}{}
 	\If {baseline heuristic in stagnation}
		\State \textbf{return} true
	\EndIf
	\State \textbf{return} false
\EndFunction
\vspace{2mm}
\Function{\texttt{in\_stagnation\_region(g)}}{}
  \If {dynamic heuristic in stagnation}
    \State \textbf{return} true
  \EndIf
  \State \textbf{return} false
\EndFunction
\vspace{2mm}
\Function{\texttt{update\_user\_guidance(arg)}}{}

\If {arg $=$ $g$}
\Comment{account for guidance}
	\If {exists suspended dynamic heuristic}
		\State {add suspended dynamic heuristic}
	\Else
		\State{get new user guidance and add dynamic heuristic}
	\EndIf
\EndIf
%
%

\If {arg $=$ $\neg g$}
\Comment{remove guidance}
  \If {\texttt{in\_stagnation\_region(g)}}
    \State {remove dynamic heuristic}
    \Comment{guidance is not useful}
  \Else   \Comment{dynamic heuristic is not in stagnation}
    \If {states passed through guidance}
      \State {discard dynamic heuristic}
      \Comment{will not be useful in future}
    \Else
      \State {suspend dynamic heuristic}
      \Comment{may be useful in future}
    \EndIf
  \EndIf
\EndIf

\EndFunction
\end{algorithmic}
\end{algorithm}

\section{Evaluation }
\label{sec:eval}
\begin{table}
 \resizebox{\linewidth}{!}{%
  \begin{tabular}{lcccc}
    \hline
    \multirow{2}{*}{} &
      \multicolumn{2}{c}{Bipedal locomotion} &
      \multicolumn{2}{c}{Ladder mounting} \\
      & (a) & (b) & (a) & (b) \\
      \hline
    Planning time(s)  & 205.2   $\pm$    22.7  & 183.4   $\pm$  25.0          & 149.4 $\pm$ 30.0   & 101.4 $\pm$ 10.3  \\
    Total time(s)     & 293.5   $\pm$    33.1  & 256.9   $\pm$  38.0          & 176.3 $\pm$ 29.6   & 128.4 $\pm$ 10.3 \\
    Expansions        & 1800.1  $\pm$    88.3  & 1770.5  $\pm$  125.9         & 980.8 $\pm$ 240.6  & 644   $\pm$ 69.3 \\
    Num. of guidances & 5.4     $\pm$    1.17  & 5.4     $\pm$  1.1           & 4.7   $\pm$ 1.2    & 5     $\pm$ 0.8   \\
    Avg. guidance time     & 16.4    $\pm$    2.8   & 13.6    $\pm$  3.6           & 6.0   $\pm$ 1.5    & 5.5   $\pm$ 0.9  \\
    \hline
  \end{tabular}} 	
  \caption{Experimental results for the bipedal locomotion and ladder mounting tasks for (a) vacillation-based detection (b) heuristic-based detection averaged over 10 trials.}
  \label{tab:stats}
	\vspace{-2mm}
\end{table}
\begin{figure*}[t]%
  \centering%
  \subfigure[]
  {
  \includegraphics[width=0.2125\textwidth]{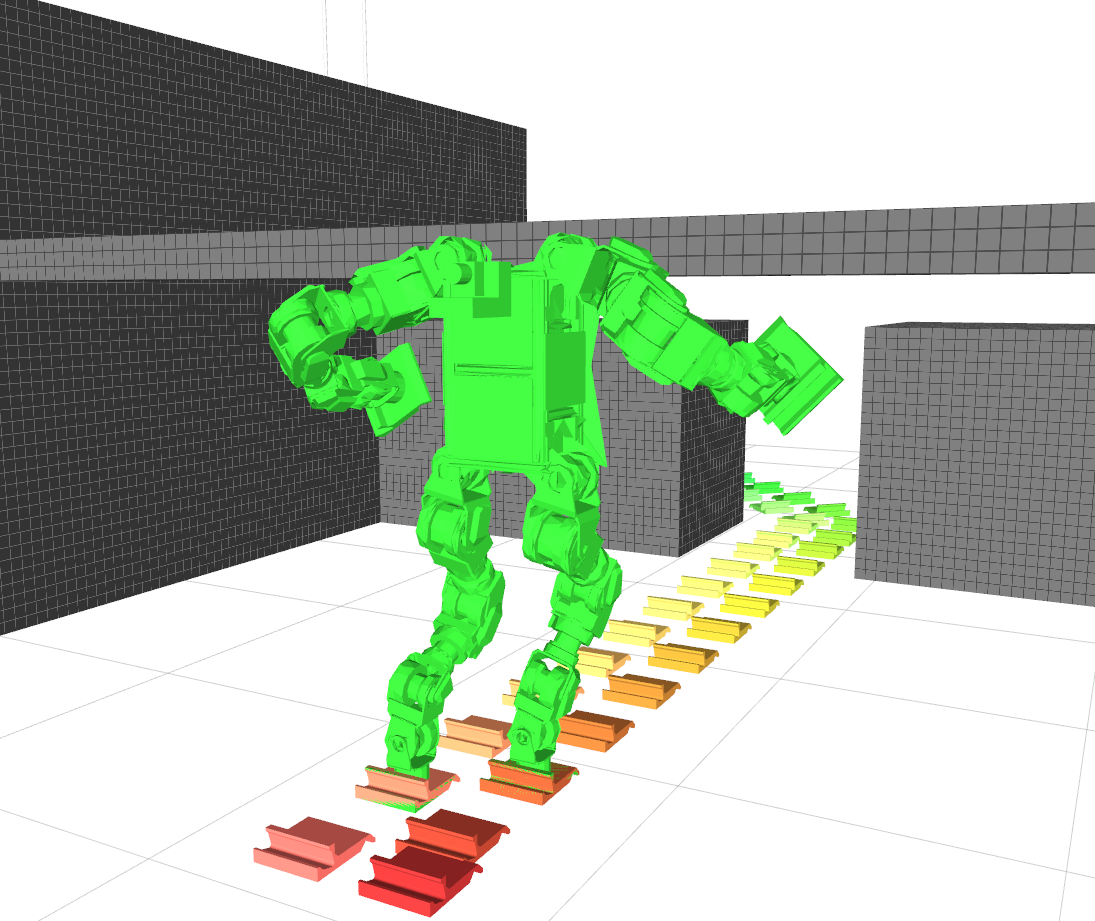}
  \label{fig:stuck_bar}
  }
  \subfigure[]
  {
  \label{fig:stuck_tables}
  \includegraphics[width=0.2125\textwidth]{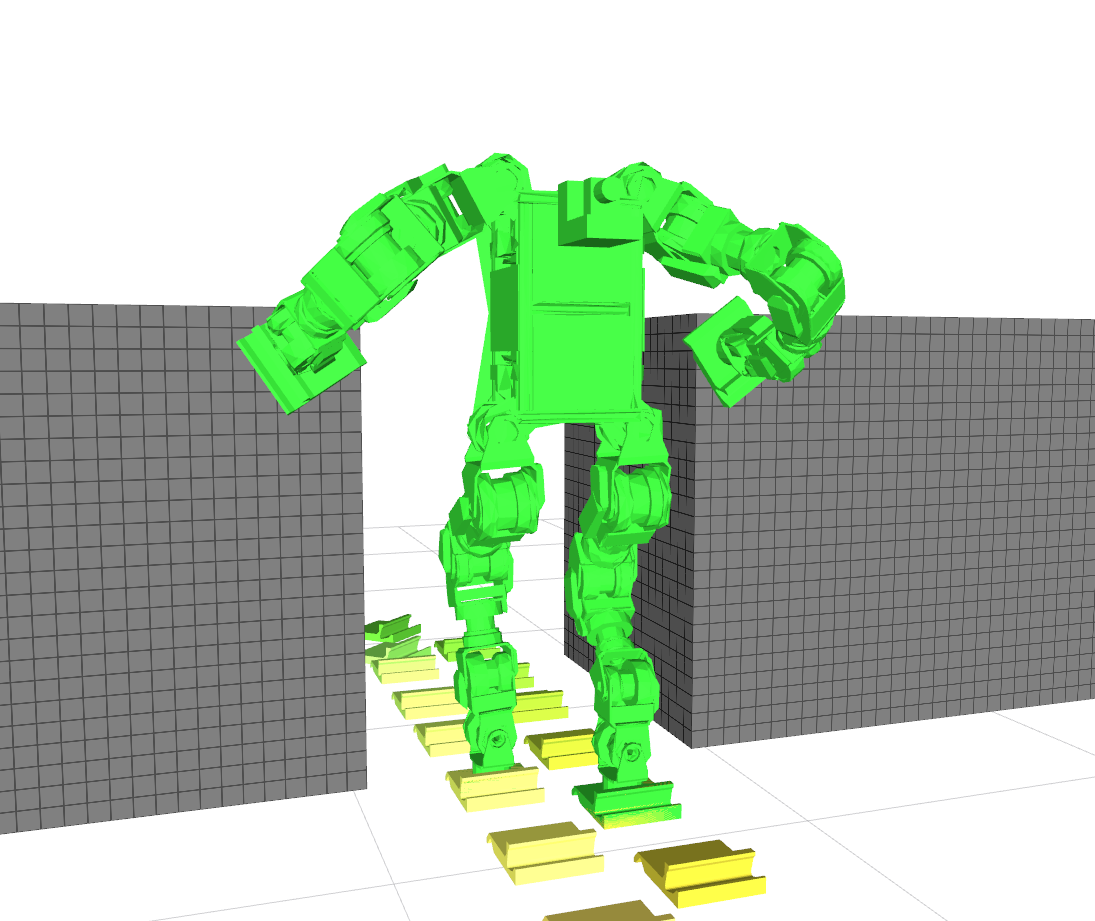}
  }
  \subfigure[]
  {
  \label{fig:stuck_door}
  \includegraphics[width=0.2125\textwidth]{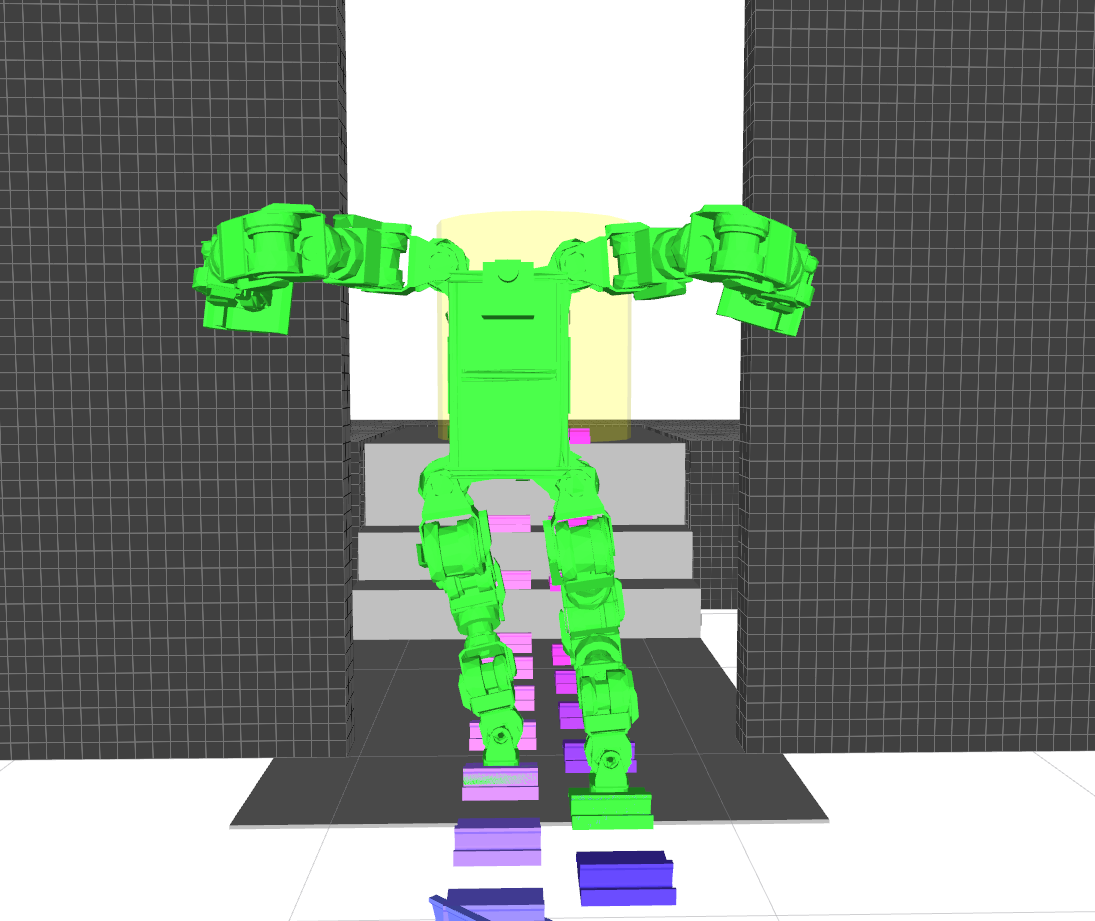}
  }  
  \subfigure[]
  {
  \label{fig:stuck_step}
  \includegraphics[width=0.2125\textwidth]{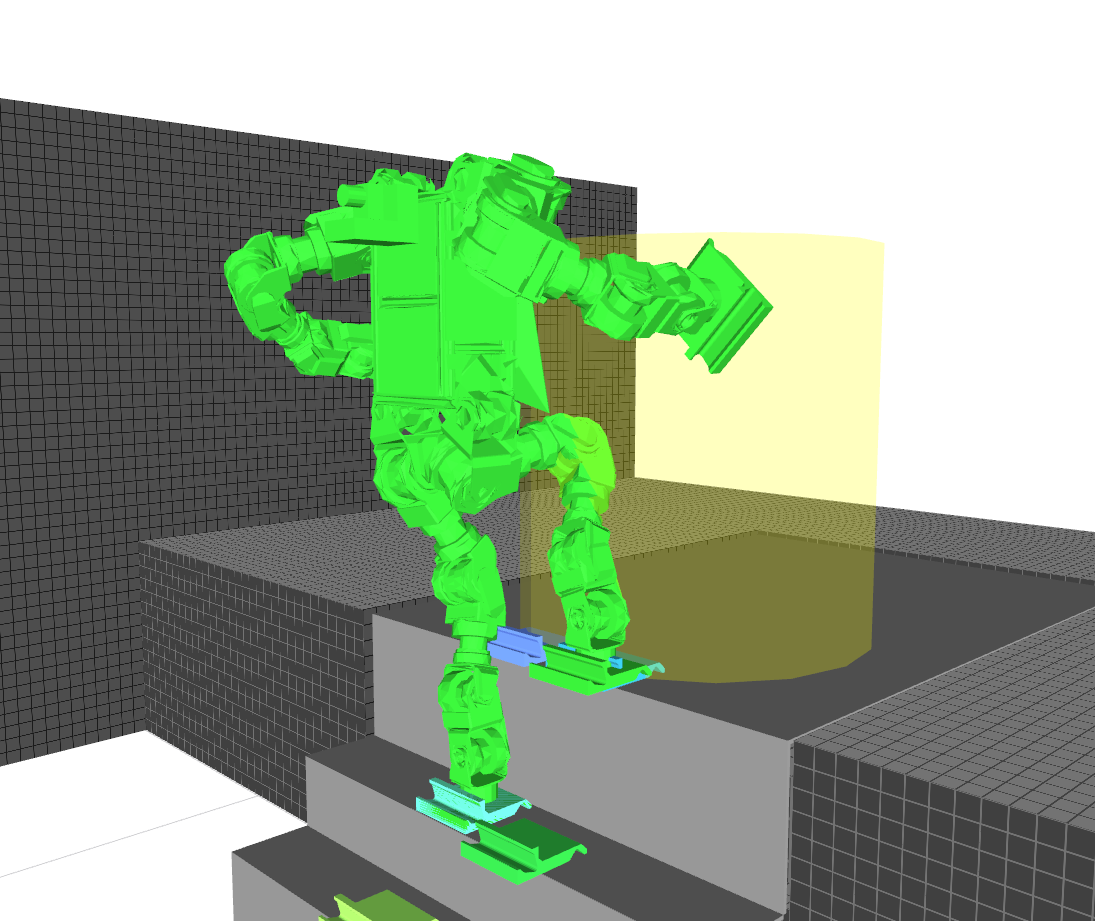}
  }
  \caption{%
    Bipedal locomotion in challenging scenarios: 
  \subref{fig:stuck_bar}~The robot has to squat or bend down to pass under a beam.
  \subref{fig:stuck_tables}~The robot has to pass through two tables by lifting its elbows high above the tables.
  \subref{fig:stuck_door}~The robot has to squeeze through a narrow doorway by tucking in its arms.
  \subref{fig:stuck_step}~The robot has to step onto a relatively high platform to reach the goal for which it has to lean more to its left side to be able to lift the right foot further up.
  }%
  \label{fig:biped_locomotion}%
  \vspace{-3.5mm}
\end{figure*}

\begin{figure}[t]%
  \centering%
  \subfigure[Vacillation-based detection]
  {
  \includegraphics[height=3.6cm]{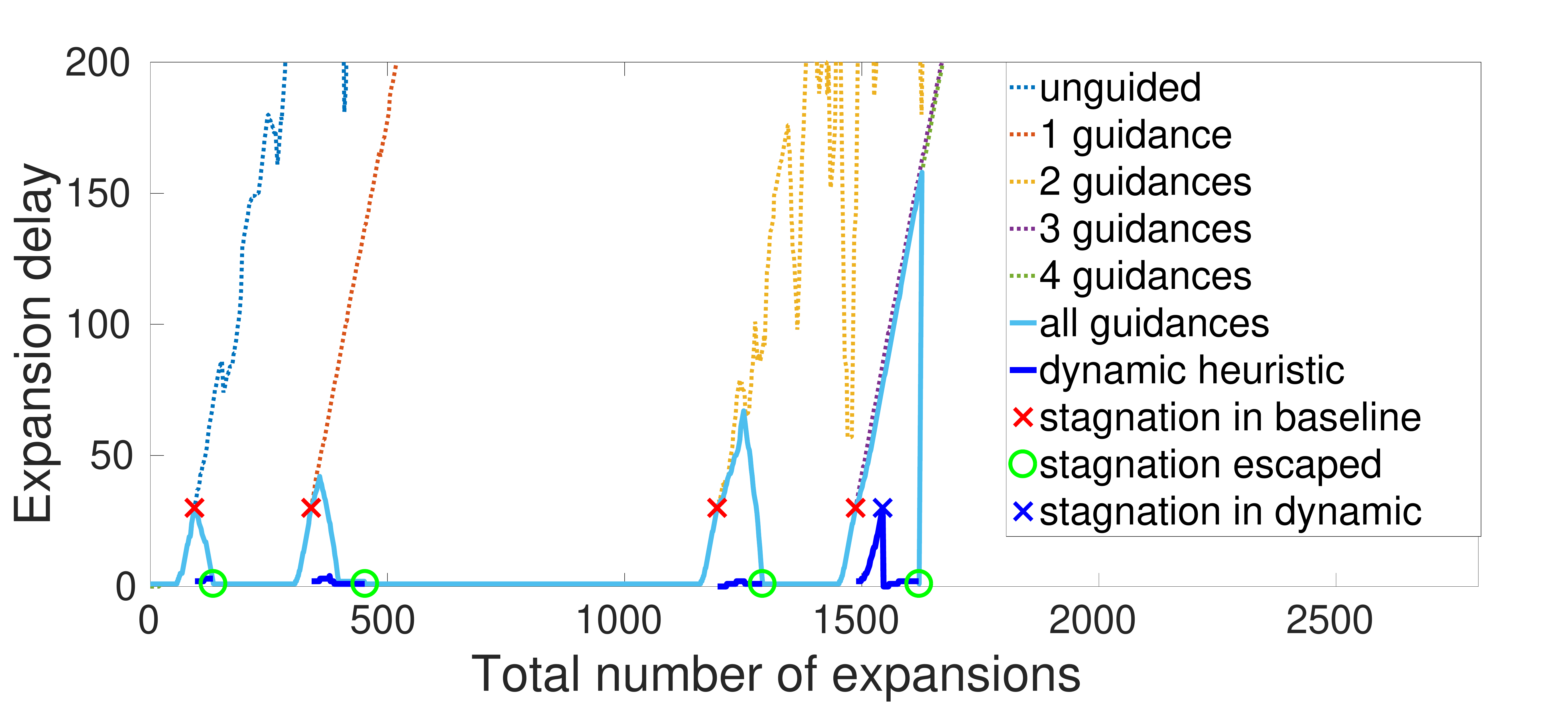}
  \label{fig:ed_plot}
  }  
	\vspace{-3mm}
  \subfigure[Heuristic-based  detection]
  {
  \includegraphics[height=3.53cm]{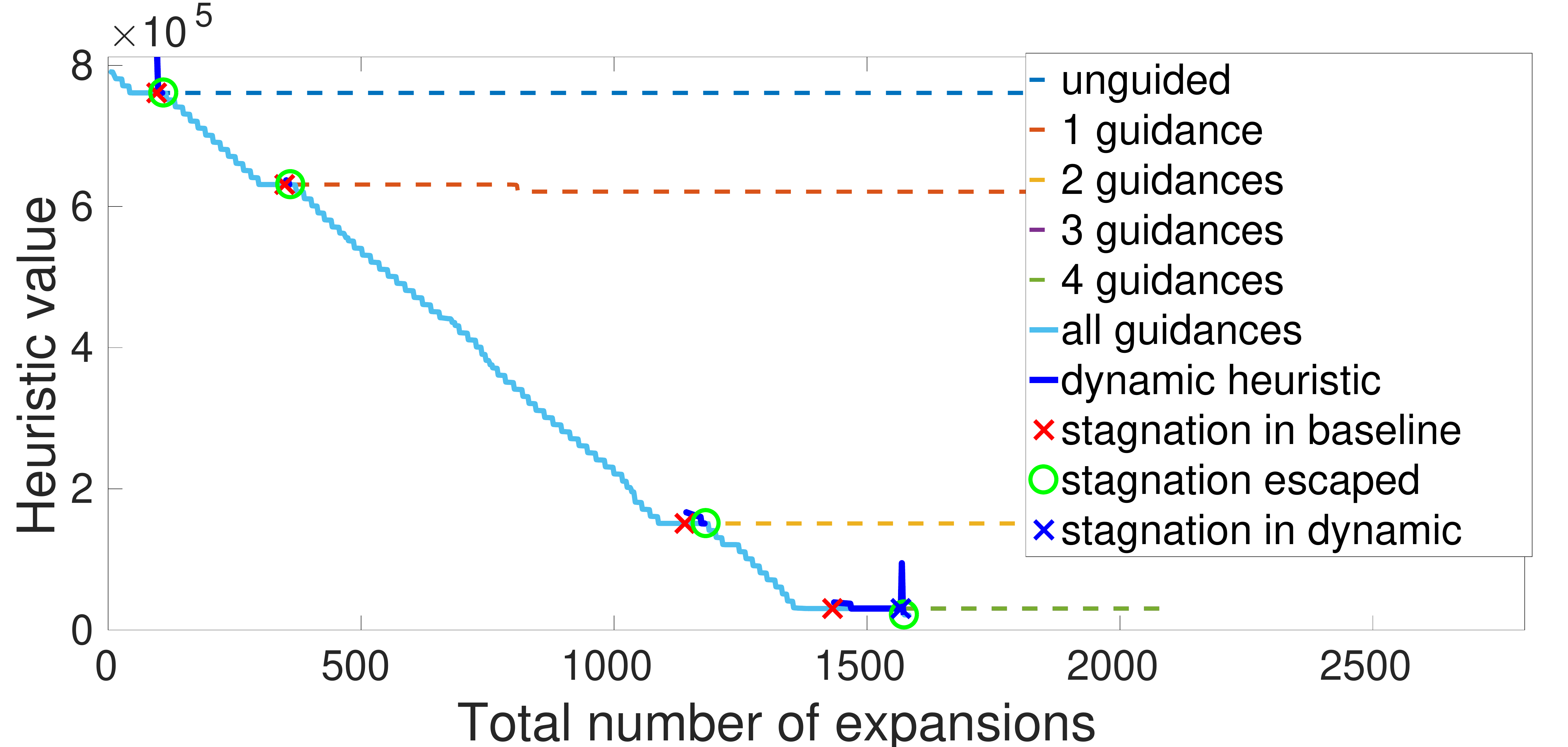}  
  \label{fig:h_plot}
  }
  \caption{%
    \mhastar progress with and without guidance.
    \subref{fig:ed_plot}~vacillation-based 
    and
    \subref{fig:h_plot}~heuristic-based
    stagnation-region detection.
		}

  \label{fig:detection_plot}%

  \vspace{-3.5mm}

\end{figure}

We evaluated the performance of our planning framework on a 34-DOF WAREC humanoid robot~\cite{MHSetal15} 
which is constrained to maintain static stability at all times. 
%
To construct the search space, we used a set of motion primitives which are short kinematically feasible motion sequences. 
These are used to generate the actions, or the successors, of a state during the search. 
For additional details see~\cite{DL18}


For each experiment we used a small set of fairly generic heuristics. 
This was done to demonstrate that our approach allows to significantly reduce the engineering effort required in carefully crafting domain-specific heuristics. 
We conducted two experiments which largely differ in the nature of mobility and thus employ different baseline heuristic functions. 
However, the dynamic heuristic function in both the experiments is the Euclidean distance in the joint 34-DOF space.
The parameter values used in both experiments are $\omega_1 (200), \omega_2 (50)$ and $\varepsilon (50)$ for the heuristic-based, and $\tau (30)$ and $\omega (10)$ for the vacillation-based stagnation-region detection.
We conducted the same experiments with a wide variety of parameters and obtained  only slight differences in the results, demonstrating that our approach is highly robust to the choice of parameters for our domain.
In addition, we conducted  similar experiments with a 7 DOF arm and with a humanoid climbing stairs while using handrails and obtained similar results to the ones we will shortly show. Those results were omitted due to lack of space.

Planning statistics, averaged over ten different trials, for all experiments are provided in Table~\ref{tab:stats}.
In all the experiments, we successfully test the hypothesis that with the help of user guidance, the planner, augmented only with a generic baseline heuristic, can completely solve challenging problems with a small number of guidances.
The deviations in the results presented in Table~\ref{tab:stats} come from the quality of the guidance provided in each run.
For the video demonstrating the approach, see \footnote{~\url{https://youtu.be/r7DHlc6XBC0}}.


\subsection{Bipedal Locomotion}
\label{subsec:locomotion}


The first task we considered is bipedal locomotion, depicted in Fig.~\ref{fig:robot}.
Here, we employ the adaptive dimensionality 
\arxiv{framework~\cite{GCBSL11,GSL12,GSL13}}{framework~\cite{GCBSL11}}
which iterates over two stages: an adaptive-planning phase which plans in a low-dimensional space when possible and a tracking phase which plans in the high-dimensional space.
Our user-guided planner is integrated within the high-dimensional planner to search for a path that tracks the footstep plan generated in the adaptive-planning  phase. 

The baseline heuristic we used was designed to assist the search with a general walking or stepping capability. For each step the footprint of the target footstep is visualized for the user to be able to better understand where the planner is stuck and how to provide the guidance. In relatively easier scenarios this heuristic would suffice in guiding the search. However if the search encounters harder situations such as those depicted in Fig.~\ref{fig:biped_locomotion}, then it is likely to get trapped into a stagnation region and would invoke user-guidance.

Results demonstrating the effectiveness of our framework are depicted in Fig.~\ref{fig:detection_plot}.
Specifically, we plot the 
method to detect stagnation regions (average expansion delay and heuristic values for vacillation-based and heuristic-based detection, respectively) as a function of the number of queue expansions.
For both methods, we ran our algorithm with and without user guidance.
Namely, if a stagnation region was detected, we recorded the state of the planner and then continued to run it once without asking the user for guidance and once using our approach. This was done every time a stagnation region was detected. 
Results show that without guidance, the planner fails to make any significant progress.
On the other hand, when guidance is given, the algorithm escapes the stagnation region and resumes to make progress towards the goal.

Interestingly, vacillation-based detection incurs a lot of noise---notice the large oscillations in the dashed yellow curve in Fig.~\ref{fig:ed_plot}. 
A possible explanation is that even in regions where the planner cannot ultimately progress towards the goal, there may be multiple short feasible paths to explore. In such settings, the expansion delay for each such path will be small while the planner does not really progress towards the goal.
Indeed, numerical results stated in Table~\ref{tab:stats} confirm that our heuristic-based method for stagnation-region detection outperforms the vacillation-based method.
We observed similar trends for the second experiment (ladder climbing) where vacillation-based method falsely detect that the stagnation region was exited (plots omitted).

\subsection{Mounting onto a Ladder}
The second task considered was mounting onto a ladder from a standing position (see Fig.~\ref{fig:gui}) to the desired contact poses. This problem is challenging because the ladder rungs and the robot grasping hooks induce narrow spaces in the configuration space which have to be traversed while maintaining stability, to establish the desired contacts. The baseline heuristic  used is the summation of four six-dimensional Euclidean distances between each of the four end effectors and their respective target poses. 
For numerical results, see Table~\ref{tab:stats}.

\section{Discussion}
\label{sec:future}

\subsubsection{Generic vs. domain-specific heuristic}
In Sec.~\ref{sec:eval} we demonstrated how user guidance allows to solve highly-constrained motion-planning problems in high-dimensional spaces with only simple baseline heuristics.
An alternative approach would be to add domain-dependent carefully-crafted heuristics~\cite{DL18,V17} which allows to faster solve the same problems completely autonomously.

When faced with a new problem domian 
(say, climbing a ladder, crawling on uneven terrain etc.)
we can either design additional heuristics to tackle this problem or leverage from user-provided online guidance.
If our problem requires planning in multiple, diverse domains this problem is accentuated---should we  use a large arsenal of heuristics that can address each domain or should we have a small set of baseline heuristics that will (roughly) address all domains and rely on user guidance when these baseline heuristics fail?
There is no clear answer to this question and our approach simply offers a general alternative to existing approaches.

\subsubsection{When ``bad'' guidance is provided}
A natural question to ask is what if the user provides a guidance that is not useful?
If the planner does not progress due to such guidance it would ask for a new one as it detects stagnation in the dynamic heuristics the same way it does in the baseline heuristic. 
As can be inferred from the experimental results presented in Sec.~\ref{sec:eval}, for the bipedal locomotion 26\% of the guidances were not useful, yet the planner was able to recover after being provided with useful guidance. 
It is important to note that while using \mhastar ensures that a path will be found regardless of the  guidance that is provided, unhelpful guidance may drive the search towards unuseful regions and waste computational resources.

\section*{Acknowledgements}
This research was in part sponsored by ARL, under the Robotics CTA program grant W911NF-10-2-0016, and by NSF Grant IIS-1409549.
%



\end{document}